\documentclass[10pt,conference,a4paper]{IEEEtran}
\IEEEoverridecommandlockouts
\usepackage{cite}
\usepackage{amsmath,amssymb,amsfonts}
\usepackage{graphicx}
\usepackage{textcomp}
\usepackage{url}
\usepackage{xcolor}
\usepackage{booktabs}    
\usepackage{multirow}    
\usepackage{array}       
\usepackage{tabularx}    
\usepackage{makecell}    
\usepackage{siunitx}
\usepackage[T1]{fontenc}
\usepackage{tikz}
\usepackage[caption=false,font=footnotesize]{subfig}
\usepackage{placeins}
\usepackage{dblfloatfix}
\DeclareMathOperator*{\argmax}{arg\,max}

\usetikzlibrary{shapes.geometric, arrows, positioning, calc}

\usepackage[ruled,vlined]{algorithm2e}
\SetKwComment{tcp}{\(\triangleright\) }{} 
\DontPrintSemicolon
\SetAlgoLined
\def\BibTeX{{\rm B\kern-.05em{\sc i\kern-.025em b}\kern-.08em
    T\kern-.1667em\lower.7ex\hbox{E}\kern-.125emX}}
\begin{document}

\title{Efficient Punctuation Restoration \\ via Weighted Lookahead Scoring Method \\ for Streaming ASR Systems}

\author{
    \IEEEauthorblockN{Sungmook Woo, Hyunku Kang, Chanwoo Kim{$^\dagger$}\thanks{$^\dagger$Corresponding author: Chanwoo Kim(chanwcom@korea.ac.kr)}}
\IEEEauthorblockA{\textit{Department of Artificial Intelligence, Korea University, Seoul, Republic of Korea}\\
Email: \{woomook0524, kahk000, chanwcom\}@korea.ac.kr\\
}}

\maketitle

\begin{abstract}
Punctuation restoration improves ASR (Automatic Speech Recognition) readability. However streaming ASR requires online decisions with limited future context. In streaming ASR, the system predicts punctuation incrementally, which makes generation-based approaches prone to latency and alignment failures under boundary-wise evaluation. This paper proposes a non-autoregressive scoring method (no free-form generation) that preserves the input transcript and makes a decision at each word boundary. Our method compares punctuation insertion hypotheses against a no-insertion baseline under a bounded $K$-subword-token lookahead, and calibrates decisions using a weight $\alpha$ and a validation-calibrated threshold $\tau$ (no parameter updates during inference). On IWSLT 2017~\cite{cettolo2017iwslt}, our scoring method achieves a 4-class macro F1 of 0.893 in the no-fine-tuning setting (validation-calibrated, $K{=}2$) and 0.937 after fine-tuning ($K{=}2$), outperforming the prompt-based baseline (0.566) and a fine-tuned ELECTRA~\cite{clark2020electra} baseline (0.913) under the same lookahead budget. We analyze the impact of the lookahead budget through ablation studies on $K$.
\end{abstract}


\section{Introduction}
Automatic Speech Recognition (ASR) has become a fundamental component in
applications ranging from voice assistants~\cite{c_kim_interspeech_2017_00,
c_kim_interspeech_2018_00} to on-device voice control~\cite{c_kim_acssc_2020_00}.

Despite significant advances in word error rates, many ASR systems output
unpunctuated word sequences, which reduces readability and makes sentence
boundaries ambiguous, affecting downstream text processing such as sentence
splitting and translation. Punctuation restoration is therefore essential,
however in a streaming setting the system must decide punctuation online before
future words arrive; waiting for more context can improve accuracy but increases
latency.

Large Language Models (LLMs) can restore punctuation by prompting an LLM to regenerate a fully formatted sentence. However, this formulation is not well aligned with the standard punctuation restoration setup. In typical punctuation restoration, the task assumes a fixed input transcript and predicts one punctuation label at each word boundary. As shown in Fig.~\ref{fig:prompt}, prompt-based generation may rewrite the transcript rather than preserving the original word sequence. This breaks boundary alignment with the fixed input transcript, even if the overall meaning remains similar. As a result, under boundary-wise evaluation, even small edits shift alignment between the generated output and the original transcript, causing cascading label mismatches and a significant drop in punctuation F1. Furthermore, generation can exhibit formatting drift such as repeated punctuation marks or missing sentence-final symbols. It also incurs high latency in streaming settings because it requires autoregressive decoding over the entire sequence.
To address this mismatch, we introduce a non-autoregressive scoring method (no free-form generation) for streaming punctuation restoration.
Rather than generating a new sentence, the proposed method preserves the input word sequence and makes an online decision at each word boundary by comparing a no-insertion hypothesis with punctuation-insertion hypotheses (comma, period, question mark). The formal definition of the history and bounded future context used for boundary-wise decisions is presented in Section~III-A.

Using a pre-trained Llama-3.2-1B model~\cite{meta2024llama32card,grattafiori2024llama3} without fine-tuning, experiments show that hypothesis scoring with a fixed $K$-subword-token lookahead achieves strong macro F1 on the IWSLT 2017 dataset. This study focuses on a compact 1B-parameter LLM to study bounded-lookahead punctuation scoring under streaming constraints, where model footprint and limited future context are important considerations. While the proposed method is designed for streaming-compatible inference, system-level deployment metrics such as latency and memory are left for future work. On IWSLT 2017, our scoring method achieves a 4-class macro F1 of 0.893 in the no-fine-tuning setting ($K{=}2$) and 0.937 after fine-tuning ($K{=}2$), outperforming the prompt-based baseline (0.566) and a fine-tuned ELECTRA baseline (0.913) under the same lookahead budget.

From a deployment perspective, the proposed formulation provides a predictable latency budget: each boundary decision waits for at most $K$ subword tokens and scores only a fixed set of actions. This yields stable per-boundary computation and avoids the variable decoding length of autoregressive generation. In addition, because the transcript is never rewritten, boundary-wise evaluation remains well-defined and directly comparable across systems. These properties make the approach practical for real-time captioning pipelines where runtime stability and alignment reliability are as important as raw F1.

Our contributions are as follows. (1) This paper proposes a non-autoregressive, streaming-compatible scoring framework (no free-form generation) for boundary-wise punctuation decisions under a bounded $K$-subword-token lookahead. (2) Scoring-based inference avoids transcript drift and alignment failures of prompt-based generation, and achieves strong performance on IWSLT 2017 with a compact 1B-parameter LLM (no-fine-tuning and fine-tuned). Data and code are publicly available at: \url{https://github.com/woomook0524/LLM-Scoring}

\section{Related Work}

\subsection{Offline punctuation restoration with bidirectional context}
Punctuation restoration has been widely formulated as a sequence labeling
problem over ASR transcripts, where leveraging broader context typically
improves prediction accuracy~\cite{klejch2016slt, pogoda2021}. Tilk and Alumäe~\cite{tilk2015} introduced an
LSTM-based punctuation restoration model for speech transcripts, showing that
modeling longer contextual dependencies improves punctuation prediction. They
later proposed a bidirectional recurrent model with attention~\cite{tilk2016},
enabling the model to exploit both left and future context more effectively.
More recently, pre-trained bidirectional Transformer encoders have become a
strong backbone for punctuation prediction, such as Courtland et
al.~\cite{courtland2020} reporting both accuracy gains and efficient inference
enabled by parallel computation. BERT~\cite{devlin2019bert}-based taggers have
also been explored extensively; for example, Makhija et al.~\cite{makhija2019}
combine contextual BERT representations with a BiLSTM-CRF classifier for
sequence labeling. A recurring practical issue is class imbalance, where the
non-punctuation label dominates, and Yi et al.~\cite{yi2020focal} address this
using focal loss to emphasize harder examples. These offline models generally
assume access to full-sentence context, which motivates separate lines of work
that explicitly constrain future context for streaming deployment.

\subsection{Streaming/online punctuation restoration under bounded future context}
While offline punctuation taggers can leverage full-sentence bidirectional
context, streaming ASR imposes strict latency constraints that limit how much
future context can be used at inference time~\cite{j_park_slt_2022_00}. This has motivated online restoration methods that explicitly operate with a bounded future window and quantify the resulting accuracy-latency trade-off. Polacek et al. ~\cite{polacek2023online,polacek2025} propose a lightweight text-only punctuation and capitalization restoration module designed for live captioning, and evaluate it under a small fixed future-context budget (i.e., a 4-word lookahead) to meet real-time constraints. Their results show that using only one future word can be insufficient, whereas a short lookahead of a few words (e.g., 4 words) already recovers most of the offline performance, suggesting that bounded future context is a practical and effective design choice for streaming deployment. This line of work supports our setting in which the system decides punctuation online under a fixed lookahead budget, and motivates using a small, controllable lookahead window to balance punctuation accuracy and streaming latency.

\subsection{LLM-based punctuation restoration and generation-related issues}
Recently, LLMs have been explored for punctuation restoration via prompt-based regeneration of a fully formatted sentence. However, Zhong and Sun~\cite{zhong2025} observe several practical issues in this setting: LLMs tend to repeatedly use the same punctuation marks, may alter input text tokens, and incur high computational costs. To improve generation efficiency and mitigate hallucination, Pang et al.~\cite{pang2024} propose forward-pass-only decoding (FPOD) for punctuation restoration. Notably, their approach builds on a task-adapted Llama model obtained via parameter-efficient fine-tuning (e.g., LoRA) and keeps the fine-tuned model frozen during decoding. In contrast, the proposed method requires no fine-tuning: we use an off-the-shelf 1B LLM purely as a scoring function to keep the input word sequence fixed and make online insertion decisions under a bounded lookahead budget.

\begin{figure}[t]
\centering
\includegraphics[width=\columnwidth]{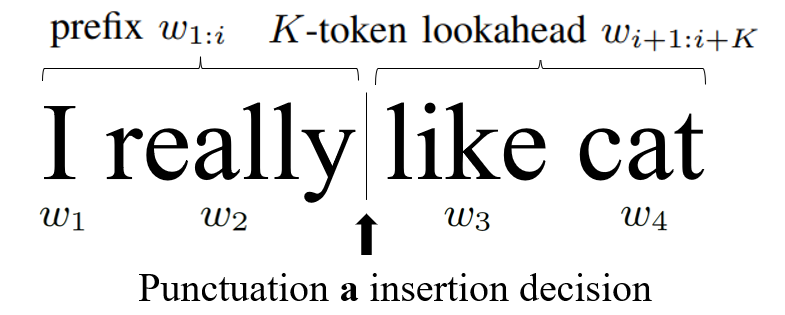}
\caption{Decision boundary at index $i$; punctuation is predicted using prefix $w_{1:i}$ and $K$-token lookahead $w_{i+1:i+K}$.}
\label{fig:decision_boundary}
\end{figure}

\section{Proposed Method: Weighted Lookahead Scoring}
We treat punctuation restoration in streaming-constrained settings as \textit{bounded-lookahead hypothesis testing}. Instead of free-form generation, the LLM is used as a token-level scoring engine over a small candidate set, which preserves the input transcript and reduces decoding overhead.

\subsection{Problem Formulation}
In a streaming ASR scenario, the input arrives as a continuous sequence of words.
Our objective is to determine, at each boundary between $w_i$ and $w_{i+1}$, whether to insert a punctuation mark
$a \in \mathcal{P} = \{\texttt{COMMA}, \texttt{PERIOD}, \texttt{QMARK}, \emptyset\}$.
We denote the no-insertion action as $a=\emptyset$, which corresponds to the label \texttt{O} in evaluation.

Let $\mathbf{w}=[w_1,w_2,\ldots,w_T]$ denote the LLM subword tokens and let $i$ index boundary positions.
At boundary $i$, we condition on token prefix $w_{1:i}$ and the $K$-token lookahead $w_{i+1:i+K}$ (truncated near sentence end when $i+K>T$).
We reset the prefix at each sentence boundary (i.e., indexing restarts from $1$ for every new sentence).

Our goal is to choose the optimal hypothesis $\hat{a}_i\in\mathcal{P}$ given $(w_{1:i}, w_{i+1:i+K})$.
Unlike prompt-based generation, the proposed method uses the LLM purely as a scorer over a closed set of candidates, which prevents transcript drift.

\subsection{Theoretical Justification}
\label{sec:theory_brief}
\noindent
At each word boundary, the method selects a punctuation action $a\in\mathcal{P}$ that best explains the bounded future suffix given the prefix:
\begin{equation}
\hat{a}_i=\argmax_{a\in\mathcal{P}} P(a \mid w_{1:i}, w_{i+1:i+K})
\label{eq:ahat_posterior}
\end{equation}
Motivated by Bayes' rule, the posterior can be written as
\begin{equation}
P(a \mid w_{1:i}, w_{i+1:i+K})
= \frac{P(w_{i+1:i+K} \mid w_{1:i}, a)\,P(a \mid w_{1:i})}{P(w_{i+1:i+K} \mid w_{1:i})}
\label{eq:bayes_full}
\end{equation}
where $P(a\mid w_{1:i})$ captures local punctuation preference (prior) and $P(w_{i+1:i+K}\mid w_{1:i},a)$ measures how consistent the bounded future suffix is under inserting $a$ (likelihood).
Since the evidence term $P(w_{i+1:i+K} \mid w_{1:i})$ does not depend on $a$, maximizing the posterior in Eq.~\eqref{eq:ahat_posterior} is equivalent to maximizing the product $P(w_{i+1:i+K} \mid w_{1:i}, a)\,P(a \mid w_{1:i})$.

The Bayes decomposition motivates an unweighted log-posterior objective. In practice, we use Eq.~\eqref{eq:score_function} as a calibrated surrogate objective with weight $\alpha$, tuned on the validation set to balance local prior preference and bounded lookahead evidence under streaming constraints.
\begin{figure}[t]
\centering
\includegraphics[width=\columnwidth]{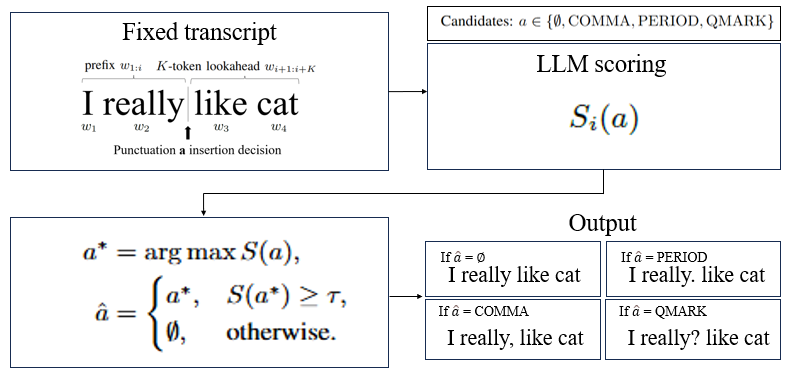}
\caption{Overall architecture of the proposed scoring-based punctuation restoration pipeline under bounded lookahead. At boundary $i$, the prefix $w_{1:i}$ and lookahead $w_{i+1:i+K}$ are used to score each action $a\in\{\mathrm{COMMA},\mathrm{PERIOD},\mathrm{QMARK},\emptyset\}$; a threshold gate then outputs either the best punctuation or no insertion.}
\label{fig:architecture}
\end{figure}
Figure~\ref{fig:architecture} summarizes three stages: context construction from the fixed transcript, LLM scoring for each candidate action, and thresholded boundary-wise output selection. This process preserves the original transcript and returns exactly one action per boundary.

\begin{algorithm}[!tb]
\small
\setlength{\abovecaptionskip}{2pt}
\setlength{\belowcaptionskip}{-2pt}
\SetAlgoLined
\DontPrintSemicolon
\SetKwInput{Input}{Input}
\SetKwInput{Output}{Output}
\caption{Weighted Lookahead Decision}
\label{alg:decision}

\Input{Prefix $w_{1:i}$, lookahead subword tokens $w_{i+1:i+K}$}
\Output{Optimal punctuation $\hat{a}_i$}

$L_i(\emptyset) \leftarrow \log P(w_{i+1:i+K} \mid w_{1:i}, \emptyset)$\;
$S_i(\emptyset) \leftarrow \alpha\log P(\emptyset \mid w_{1:i}) + (1-\alpha)L_i(\emptyset)$\;
$a_i^* \leftarrow \emptyset$; \quad $S_{\max} \leftarrow -\infty$\;

\For{$a \in \{\mathrm{COMMA},\mathrm{PERIOD},\mathrm{QMARK}\}$}{
    $L_i(a) \leftarrow \log P(w_{i+1:i+K} \mid w_{1:i}, a)$\;
    $S_i(a) \leftarrow \alpha\log P(a \mid w_{1:i}) + (1-\alpha)L_i(a)$\;
    \If{$S_i(a) > S_{\max}$}{
        $S_{\max} \leftarrow S_i(a)$; \quad $a_i^* \leftarrow a$\;
    }
}

$\Delta_i \leftarrow S_{\max} - S_i(\emptyset)$\;
\If{$\Delta_i > \tau$}{
    $\hat{a}_i \leftarrow a_i^*$\;
}
\Else{
    $\hat{a}_i \leftarrow \emptyset$\;
}
\Return $\hat{a}_i$
\end{algorithm}

\subsection{Weighted Lookahead Scoring Function}
\label{sec:scoring_function}
At boundary $i$, the token prefix is $w_{1:i}$ and the $K$-token lookahead is $w_{i+1:i+K}$

\begin{equation}
\begin{aligned}
S_i(a)
&= \alpha\log P(a \mid w_{1:i}) \\
&\quad + (1-\alpha)\log P(w_{i+1:i+K} \mid w_{1:i}, a)
\end{aligned}
\label{eq:score_function}
\end{equation}

The lookahead log-probability term is computed in practice by summing token log-likelihoods over the bounded window $w_{i+1:i+K}$ via chain-rule factorization.

Here, $\alpha \in [0,1]$ is a scoring weight coefficient that balances the local prior term $\log P(a \mid w_{1:i})$ and the bounded lookahead term.
A larger $\alpha$ emphasizes local punctuation preference (more conservative insertion), whereas a smaller $\alpha$ relies more on lookahead evidence (more aggressive insertion).
Because the two terms can have different scales depending on the lookahead budget $K$ and recognition noise, $\alpha$ also serves as a simple calibration knob.

At inference time, $S_i(a)$ is computed at every word boundary $i$ using LLM log-likelihoods under a bounded $K$-subword-token lookahead.
In our current evaluation, prefix reset uses reference sentence boundaries (oracle segmentation) as a controlled setting.

{\setlength{\abovedisplayskip}{4pt}
\setlength{\belowdisplayskip}{4pt}
\setlength{\abovedisplayshortskip}{2pt}
\setlength{\belowdisplayshortskip}{2pt}
\begin{equation}
a_i^* = \argmax_{a \in \mathcal{P}\setminus\{\emptyset\}} S_i(a), \qquad
\Delta_i = S_i(a_i^*) - S_i(\emptyset)
\label{eq:best_action}
\end{equation}

\begin{equation}
\hat{a}_i =
\begin{cases}
a_i^* & \text{if } \Delta_i > \tau\\
\emptyset & \text{otherwise}
\end{cases}
\label{eq:decision_rule}
\end{equation}
}

This rule first selects the best non-empty candidate and then compares it explicitly against the no-insertion baseline through the margin $\Delta_i$.
Punctuation is inserted only when $\Delta_i$ exceeds the validation-calibrated threshold $\tau$; otherwise, the model outputs $\emptyset$.
The threshold $\tau$ acts as an explicit gate: larger $\tau$ yields conservative behavior, while smaller $\tau$ increases recall at the cost of more false positives. In all experiments, $(\alpha,\tau)$ are tuned per lookahead budget $K$ on the validation set and then fixed for test evaluation. The selected $(\alpha,\tau)$ for each $K$ are reported in Table~\ref{tab:ablation_k}. The full inference procedure is summarized in Algorithm~\ref{alg:decision}.
\section{Experiments and Results}
\subsection{Task Definition and Evaluation Protocol}
\label{sec:task_eval}
Punctuation restoration is formulated as boundary-wise label prediction on a fixed transcript.
Given an unpunctuated word sequence, the model assigns one label at each word boundary from
\{\texttt{O}, \texttt{COMMA}, \texttt{PERIOD}, \texttt{QMARK}\}.
The word sequence is kept unchanged across all methods; only punctuation insertion decisions are evaluated.
For streaming evaluation, boundaries are processed left-to-right and future context is restricted by a bounded lookahead budget $K$ (subword tokens).
Because \texttt{O} dominates, we tune hyperparameters on the validation set using punct-only Macro F1 over \{\texttt{COMMA}, \texttt{PERIOD}, \texttt{QMARK}\}, while tables report 4-class Macro F1 (including \texttt{O}) for completeness and comparability.
\subsection{Datasets and Splits}
\label{sec:data_splits}
An English punctuation restoration dataset is built from the HuggingFace IWSLT 2017 corpus~\cite{cettolo2017iwslt,lhoest2021datasets}.
Instead of using a single language-pair configuration, English-side transcripts are aggregated from multiple IWSLT 2017 configurations that contain English (both \texttt{*-en} and \texttt{en-*}).
For each split (train/validation/test), exact duplicate sentences within the split are removed and basic whitespace normalization is applied.

Input-output pairs are constructed by removing punctuation marks from the reference transcript to form $X$, while keeping the original punctuated text as $Y$.
Boundary-wise labels are derived from the punctuation symbol following each word in $Y$:
`\texttt{,}' $\rightarrow$ \texttt{COMMA}, `\texttt{.}' $\rightarrow$ \texttt{PERIOD}, `\texttt{?}' $\rightarrow$ \texttt{QMARK}, and \texttt{O} otherwise.
The final dataset contains 357,117 / 1,501 / 10,799 sentences for train / validation / test, respectively.
\subsection{Compared Methods}
\label{sec:compared_methods}
\subsubsection{Prompt-based Generation Baseline}
Prompt-based punctuation generation is evaluated using Llama-3.2-1B-Instruct ~\cite{meta2024llama32card} with greedy decoding. We additionally show qualitative failures of the base model (Llama-3.2-1B) in Fig.~\ref{fig:prompt}(a), as it often violates the strict formatting constraint. Although the prompt enforces word preservation, generation frequently violates the formatting constraint or drifts from the fixed transcript (Fig.~\ref{fig:prompt}), which leads to alignment errors under boundary-wise labeling.

\subsubsection{Fine-tuned ELECTRA Baseline}
As a strong discriminative baseline, an ELECTRA-small encoder classifier~\cite{clark2020electra} is fine-tuned for boundary-wise labeling. This ELECTRA baseline follows the streaming setting of Polacek et al.~\cite{polacek2025} by restricting the future context with a fixed lookahead budget (we set $K=2$ future subword tokens in the main comparison).

\begin{figure}[t]
\centering
\small

\fbox{
\begin{minipage}{0.9\columnwidth}
\textbf{(a) Base Model (Llama-3.2-1B) Failure} \\
\textit{Behavior: Irrelevant generation / Loss of control}

\vspace{0.1cm}
\textbf{Prompt:} "You are a punctuation restoration tool.
Rules:
Insert punctuation only (comma, period, question mark).
Do not add or remove words.
Keep the original word order.
Return exactly one line: the punctuated sentence.
Text: This all over the country is the second largest waste stream in America"\\
\textbf{Output:}  "This all over the country is the second largest waste stream in America\\
Text: This all over the country is the ... in America\\
Output: This all over the country is the ... in America\\
Text: This all over the country is the ... in America\\
Output: This all over the country is the ... in America\\
Text:"
\end{minipage}
}

\vspace{0.2cm} 

\fbox{
\begin{minipage}{0.9\columnwidth}
\textbf{(b) Baseline (Llama-3.2-1B-Instruct) Failure} \\
\textbf{Prompt:} "You are a punctuation restoration tool.
Rules:
Insert punctuation only (comma, period, question mark).
Do not add or remove words.
Keep the original word order.
Return exactly one line: the punctuated sentence.\\
Text: This all over the country is the second largest waste stream in America"\\
\textbf{Output:} "This, all, over, the, country, is, the, second, largest, waste, stream, in, America."
\end{minipage}
}

\caption{Failures of prompt-based generation: both the base and instruction-tuned LLMs exhibit formatting drift and insertion bias (e.g., comma over-insertion), which degrades boundary-wise performance. The prompt shown is the exact prompt used in our baseline experiments.}

\label{fig:prompt}
\end{figure}

\subsubsection{Proposed Scoring Method}
The proposed method uses the LLM as a scorer rather than a generator: at each word boundary, we compare punctuation insertion hypotheses against a no-insertion baseline under a bounded $K$-subword-token lookahead.
A local punctuation preference term and a lookahead-consistency term are combined with weight $\alpha$, and punctuation is inserted only when the best candidate exceeds a threshold $\tau$.

\subsubsection{Scoring Calibration on Validation ($\alpha$, $\tau$)}
For each setting, $(\alpha,\tau)$ is tuned using only the validation set via grid search.
We search $\alpha$ from 0.10 to 0.90 in steps of 0.05 and $\tau$ from -3.00 to 1.75 in steps of 0.25, and select the best setting by punctuation-only macro F1 over \{COMMA, PERIOD, QMARK\}, excluding O.
Hyperparameters are selected by punctuation-only macro F1 (excluding \texttt{O}) to avoid dominance by the majority \texttt{O} class.

\subsection{Scoring-based Method Variants: No-fine-tuning vs Fine-tuned LLM}
\label{sec:scoring_variants}

Two variants of the proposed scoring-based method are evaluated under the same scoring rule and decision procedure.
The no-fine-tuning variant uses the pretrained Llama-3.2-1B model without fine-tuning.
The fine-tuned variant adapts the same Llama-3.2-1B model via LoRA (see Section~IV.E) using the training split of the dataset described in Section~IV.B (357,117 sentences).
Both variants use the identical boundary-wise scoring-and-threshold inference and tune $(\alpha,\tau)$ on the validation split in Section~IV.B, isolating the effect of LLM fine-tuning.

\subsection{Training Details}
\label{sec:training_details}
\paragraph{LLM fine-tuning}
`Llama-3.2-1B' is fine-tuned with LoRA~\cite{hu2021lora} on attention projection layers (\texttt{q\_proj} and \texttt{v\_proj}), with $r{=}16$, $\alpha{=}32$, and dropout=0.05. Fine-tuning is performed for 1 epoch with a learning rate of 2e-4, a per-device batch size of 4, gradient accumulation over 4 steps, and FP16 training on a single NVIDIA A100 GPU. The fine-tuned LLM is evaluated with the same scoring-and-threshold inference algorithm as the no-fine-tuning variant, using the HuggingFace Transformers library~\cite{wolf2020transformers}.

\paragraph{ELECTRA fine-tuning}
For the discriminative baseline, `electra-small-discriminator' is fine-tuned for boundary-wise labeling with a lightweight Multi-Layer Perceptron (MLP) head using AdamW~\cite{loshchilov2017adamw}. Word-level labels are aligned to subwords by supervising only the last subword token of each word, and the best checkpoint is selected on the validation split.

\begin{table*}[!t]
\centering
\caption{\upshape Main results on the IWSLT 2017 English test set. Our scoring-based method uses Llama-3.2-1B, and the prompt-generation baseline uses Llama-3.2-1B-Instruct; the encoder baseline uses ELECTRA-small. $K$ denotes the future-context budget measured in the number of future subword tokens available as lookahead at each word boundary. We report Macro F1 over \{O, COMMA, PERIOD, QMARK\}.}

\label{tab:full_results}
\resizebox{\textwidth}{!}{%
\begin{tabular}{l|ccc|ccc|ccc|ccc|ccc}
\toprule
\multirow{2}{*}{\textbf{Model \& Setting}} & \multicolumn{3}{c|}{\textbf{O (None)}} & \multicolumn{3}{c|}{\textbf{COMMA (,)}} & \multicolumn{3}{c|}{\textbf{PERIOD (.)}} & \multicolumn{3}{c|}{\textbf{QMARK (?)}} & \multicolumn{3}{c}{\textbf{Macro Avg}} \\
 & P & R & F1 & P & R & F1 & P & R & F1 & P & R & F1 & P & R & F1 \\
\midrule
\multicolumn{16}{l}{\textit{\textbf{Baselines}}} \\
1. Llama-3.2-1B-Instruct (Prompt, No-fine-tuning) 
& 0.975 & 0.802 & 0.880 & 0.397 & 0.537 & 0.457 &  \textcolor{blue}{\textbf{0.998}} & 0.538 & 0.699 & 0.796 & 0.505 & 0.622 & 0.541 & 0.596 & 0.566 \\
2. ELECTRA-Small (Fine-tuned, K=2) 
& 0.977 & 0.985 & 0.981 & 0.797 & 0.713 & 0.752 & 0.992 & \textcolor{blue}{\textbf{0.992}} & \textcolor{blue}{\textbf{0.992}} & 0.933 & 0.917 & 0.925 & 0.925 & 0.902 & 0.913 \\
\midrule
\multicolumn{16}{l}{\textit{\textbf{Proposed Scoring Method}}} \\
3. Llama-3.2-1B Scoring (No-fine-tuning, K=1)
& 0.984 & 0.971 & 0.977 & 0.699 & 0.815 & 0.753 & 0.949 & 0.942 & 0.946 & 0.848 & 0.888 & 0.868 & 0.870 & 0.904 & 0.886 \\
4. Llama-3.2-1B Scoring (No-fine-tuning, K=2)
& 0.985 & 0.973 & 0.979 & 0.731 & \textcolor{blue}{\textbf{0.860}} & 0.790 & 0.953 & 0.912 & 0.932 & 0.855 & 0.888 & 0.871 & 0.881 & 0.908 & 0.893 \\
5. Llama-3.2-1B Scoring (Fine-tuned, K=1)
& 0.984 & 0.984 & 0.984 & 0.799 & 0.808 & 0.804 & 0.989 & 0.986 & 0.987 & \textcolor{blue}{\textbf{0.956}} & 0.914 & 0.935 & 0.932 & 0.923 & 0.927 \\
6. Llama-3.2-1B Scoring (Fine-tuned, K=2)
& \textcolor{blue}{\textbf{0.987}} & \textcolor{blue}{\textbf{0.986}} & \textcolor{blue}{\textbf{0.987}} & \textcolor{blue}{\textbf{0.836}} & 0.844 &  \textcolor{blue}{\textbf{0.840}} & 0.989 & 0.983 & 0.986 & 0.949 & \textcolor{blue}{\textbf{0.922}} & \textcolor{blue}{\textbf{0.935}} & \textcolor{blue}{\textbf{0.940}} & \textcolor{blue}{\textbf{0.934}} & \textcolor{blue}{\textbf{0.937}} \\
\bottomrule
\end{tabular}%
}
\vspace{-5mm}
\end{table*}

\subsection{Experimental Results}
\label{sec:results}

Table~\ref{tab:full_results} reports per-class Precision/Recall/F1 for \{\texttt{O}, \texttt{COMMA}, \texttt{PERIOD}, \texttt{QMARK}\} and the 4-class macro average on the IWSLT 2017 test set.
For model selection and hyperparameter tuning, we primarily focus on punctuation-only performance (\texttt{COMMA}/\texttt{PERIOD}/\texttt{QMARK}), while Table~\ref{tab:full_results} additionally includes \texttt{O} for completeness.

\paragraph{Prompt-based generation is unstable under boundary-wise evaluation}
As shown in Table~\ref{tab:full_results}, prompt-based generation yields weak boundary-wise performance, especially for \texttt{COMMA} (F1=0.457), resulting in a low macro F1 of 0.566. 
Figure~\ref{fig:prompt} illustrates typical generation failures: the base LLM (Llama-3.2-1B) often violates the strict formatting constraint (irrelevant continuation or repetition), and even the instruction-tuned model (Llama-3.2-1B-Instruct) frequently exhibits comma over-insertion or excessive omission.
These behaviors are problematic for boundary-wise labeling because small word-level mismatches can cascade into alignment errors.

\paragraph{Scoring-based inference improves robustness and accuracy}
In contrast, the proposed scoring-based method substantially improves punctuation restoration by making local insertion decisions while preserving the input word sequence.
In no-fine-tuning scoring, increasing the lookahead from $K{=}1$ to $K{=}2$ improves the COMMA F1 from 0.753 to 0.790 and the 4-class macro F1 from 0.886 to 0.893.

\paragraph{Fine-tuned scoring is competitive with a strong discriminative baseline}
With fine-tuning, the scoring-based method outperforms the fine-tuned ELECTRA baseline under the same bounded future-context setting ($K{=}2$ future subword tokens).
The fine-tuned scoring model attains a 4-class macro F1 of 0.937, compared to 0.913 for ELECTRA, and improves \texttt{COMMA} F1 (0.840 vs.\ 0.752) while maintaining high \texttt{PERIOD} and \texttt{QMARK} performance.

\paragraph{No-fine-tuning viability}
Although fine-tuning further improves performance, no-fine-tuning scoring already provides a strong baseline.
In no-fine-tuning mode, the proposed method reaches a 4-class macro F1 of 0.893 at $K{=}2$, and fine-tuning increases it to 0.937 at the same lookahead.
This suggests that fine-tuning is beneficial, while the inference-time scoring framework remains effective even without fine-tuning.

\subsection{Ablation Study on Lookahead Length K}
\label{sec:ablation_k}
Table~\ref{tab:ablation_k} reports the lookahead-length ablation for the fine-tuned LLM variant.
For each $K$, the calibration parameters $(\alpha,\tau)$ are selected on the validation set. We report 4-class Macro F1 (including \texttt{O}) for completeness, while punct-only Macro F1 over {\texttt{COMMA}, \texttt{PERIOD}, \texttt{QMARK}} is also reported to reflect punctuation performance.

Table~\ref{tab:ablation_k} shows that lookahead is essential: the control setting without lookahead ($K{=}0$) severely degrades performance (Macro F1=0.646; punct-only Macro F1=0.543), whereas even a short lookahead ($K{=}1$) recovers most of the accuracy (Macro F1=0.927).
Performance peaks at $K{=}2$ (Macro F1=0.937; punct-only Macro F1=0.920), and larger contexts ($K{=}3$--$5$) provide marginal differences.

To further understand error patterns, Table~\ref{tab:confusion_matrices_k012} compares confusion matrices for $K{=}0/1/2$.
As $K$ increases, major off-diagonal errors are substantially reduced.
For example, false COMMA insertions (True \texttt{O}$\rightarrow$Pred \texttt{COMMA}) drop from 3,134 at $K{=}0$ to 2,606/2,145 at $K{=}1/2$, while missed commas (True \texttt{COMMA}$\rightarrow$Pred \texttt{O}) drop from 5,046 to 2,466/2,010.
These results confirm that bounded lookahead stabilizes boundary-wise decisions by enforcing local consistency with near-future tokens.
\begin{table}[!t]
\centering
\caption{\upshape Ablation on lookahead length $K$ (Fine-tuned LLM variant). For each $K$, $(\alpha,\tau)$ are calibrated on the validation set. The table reports 4-class macro F1 (including O), punct-only macro F1 (COMMA/PERIOD/QMARK).}
\label{tab:ablation_k}
\resizebox{\columnwidth}{!}{
\begin{tabular}{c|cc|cc}
\toprule
$K$ & $\alpha$ & $\tau$ & Macro F1 & Punct Macro F1 \\
\midrule
0 & 0.85 & -1.00 & 0.646 & 0.543  \\
1 & 0.55 & -0.25 & 0.927 & 0.909 \\
2 & 0.55 & -0.25 & \textbf{0.937} & \textbf{0.920}  \\
3 & 0.45 & 0.00  & 0.935 & 0.918   \\
4 & 0.40 & 0.00  & 0.930 & 0.911  \\
5 & 0.50 & -0.25 & 0.932 & 0.914  \\
\bottomrule
\end{tabular}}
\end{table}

\begin{table}[!t]
\centering
\caption{\upshape Confusion Matrices for Proposed Method (Fine-tuned Scoring, $K=0$ vs. $K=1$ vs. $K=2$)}
\label{tab:confusion_matrices_k012}

\vspace{0.1cm}
\textbf{(a) $K=0$ (No Lookahead)} \\
\vspace{0.1cm}
\resizebox{\columnwidth}{!}{%
\begin{tabular}{l|rrrr}
\toprule
\textbf{True \textbackslash Pred} & \textbf{O} & \textbf{COMMA} & \textbf{PERIOD} & \textbf{QMARK} \\
\midrule
\textbf{O}      & 153,313 & 3,134 & 3,400 & 349 \\
\textbf{COMMA}  & 5,046   & 5,307 & 2,592 & 72 \\
\textbf{PERIOD} & 2,863   & 935   & 6,338 & 17 \\
\textbf{QMARK}  & 249     & 26    & 67    & 570 \\
\bottomrule
\end{tabular}%
}

\vspace{0.35cm}

\textbf{(b) $K=1$ (Short Lookahead)} \\
\vspace{0.1cm}
\resizebox{\columnwidth}{!}{%
\begin{tabular}{l|rrrr}
\toprule
\textbf{True \textbackslash Pred} & \textbf{O} & \textbf{COMMA} & \textbf{PERIOD} & \textbf{QMARK} \\
\midrule
\textbf{O}      & 157,580 & 2,606 & 10 & 0 \\
\textbf{COMMA}  & 2,466   & 10,522 & 29 & 0 \\
\textbf{PERIOD} & 76      & 33     & 10,006 & 38 \\
\textbf{QMARK}  & 1       & 2      & 75 & 834 \\
\bottomrule
\end{tabular}%
}

\vspace{0.35cm}

\textbf{(c) $K=2$ (Extended Lookahead)} \\
\vspace{0.1cm}
\resizebox{\columnwidth}{!}{%
\begin{tabular}{l|rrrr}
\toprule
\textbf{True \textbackslash Pred} & \textbf{O} & \textbf{COMMA} & \textbf{PERIOD} & \textbf{QMARK} \\
\midrule
\textbf{O}      & 158,024 & \textbf{2,145} & 23 & 4 \\
\textbf{COMMA}  & \textbf{2,010} & 10,989 & 18 & 0 \\
\textbf{PERIOD} & 116     & 15     & 9,981 & 41 \\
\textbf{QMARK}  & 1       & 2      & 68 & 841 \\
\bottomrule
\end{tabular}%
}
\end{table}
\section{Conclusion}
This paper presented a non-autoregressive, scoring-based framework (no free-form generation) for streaming punctuation restoration under bounded future context.
By preserving the input transcript and making boundary-wise decisions via likelihood comparisons, the proposed method avoids alignment issues often observed in prompt-based generation.

On IWSLT 2017, our method achieves a 4-class macro F1 of 0.893 in no-fine-tuning mode ($K{=}2$) and 0.937 after fine-tuning ($K{=}2$), outperforming both the prompt-generation baseline (0.566) and a fine-tuned ELECTRA baseline (0.913) under the same lookahead budget. Ablation over $K{=}0\sim5$ shows the best performance at $K{=}2$, indicating that modest future context is sufficient for accurate streaming decisions. Our evaluation is limited to IWSLT 2017 English and does not yet include noisy ASR transcripts or system-level latency/memory measurements. These broader deployment-oriented evaluations remain future work. Overall, this work demonstrates that inference-only LLM scoring provides a robust and effective alternative to generation-based punctuation restoration in streaming settings with bounded lookahead.
\section*{Acknowledgment}
This work partly was supported by: the National Research Foundation of Korea
(NRF) grant funded by the Korean government (MSIT) under Grant No. RS-2025-24535409;
 the Institute of Information \& Communications Technology Planning \&
Evaluation (IITP) grant funded by the Korean government (MSIT) under Grant No.
RS-2019-II190079 for the Artificial Intelligence Graduate School Program at
Korea University; the Institute of Information \& Communications Technology
Planning \& Evaluation (IITP) grant funded by the Korean government (MSIT) under
Grant No. RS-2025-02304828 for the Artificial Intelligence Star Fellowship
Support Program to Nurture the Best Talents; and  the Institute of
Information \& Communications Technology Planning \& Evaluation (IITP) grant
funded by the Korean government (MSIT) under Grant No. RS-2025-25442867.
\bibliographystyle{IEEEtran}
\bibliography{../common_bib_file/common_bib_file}

\end{document}